\documentclass[letterpaper, 10 pt, conference]{ieeeconf}  

\IEEEoverridecommandlockouts                              

\usepackage{graphics} 
\usepackage{epsfig} 
\usepackage{xcolor}
\usepackage{hyperref}

\usepackage[T1]{fontenc}

\title{\LARGE \bf
iCub Knows Where You Look: Exploiting Social Cues for Interactive Object Detection Learning
}

\author{Maria Lombardi$^{1*}$, Elisa Maiettini$^{1*}$, Vadim Tikhanoff$^{2}$ and Lorenzo Natale$^{1}$
\thanks{*These authors contributed equally to the work}
\thanks{$^{1}$Maria Lombardi, Elisa Maiettini and Lorenzo Natale are with Humanoids, Sensing and Perception Group, Istituito Italiano di Tecnologia, Genoa, Italy.
        {\tt\small maria.lombardi1@iit.it, elisa.maiettini@iit.it, lorenzo.natale@iit.it}}%
\thanks{$^{2}$Vadim Tikhanoff is with iCubTech facility, Istituto Italiano di Tecnologia, Genoa, Italy.
        {\tt\small vadim.tikhanoff@iit.it}}%
}

\begin{document}

\maketitle
\thispagestyle{empty}
\pagestyle{empty}


\begin{abstract}
Performing joint interaction requires constant mutual monitoring of own actions and their effects on the other's behaviour. Such an action-effect monitoring is boosted by social cues and might result in an increasing sense of agency. Joint actions and joint attention are strictly correlated and both of them contribute to the formation of a precise temporal coordination.
In human-robot interaction, the robot's ability to establish joint attention with a human partner and exploit various social cues to react accordingly is a crucial step in creating communicative robots. Along the social component, an effective human-robot interaction can be seen as a new method to improve and make the robot's learning process more natural and robust for a given task.
In this work we use different social skills, such as mutual gaze, gaze following, speech and human face recognition, to develop an effective teacher-learner scenario tailored to visual object learning in dynamic environments. Experiments on the iCub robot demonstrate that the system allows the robot to learn new objects through a natural interaction with a human teacher in presence of distractors.
\end{abstract}


\section{INTRODUCTION}
\label{sec:introduction}
Acting together to fulfil a common goal requires constant mutual monitoring of each other's actions. Such a monitoring ensures that each coactor is aware of the own actions and of their effects in the other's behaviour. The perception of the causality between the action and its effect is highly linked to the emergence of a Sense of Agency (SoA). SoA refers to the feeling of being in control of one's own action and their outcomes~\cite{Gallagher2000,Haggard2017}. It is closely related to the known phenomenon of the temporal binding (TB), largely used as implicit measure of agency. The TB effect refers to the subjective and systematic underestimation of time intervals between two related events, most commonly the action and its effect~\cite{Barlas2013}. An increase in the perceived time interval between the cause and its effect leads to a reduction in SoA.

\begin{figure}[t]
    \centering
    \includegraphics[width=0.8\columnwidth]{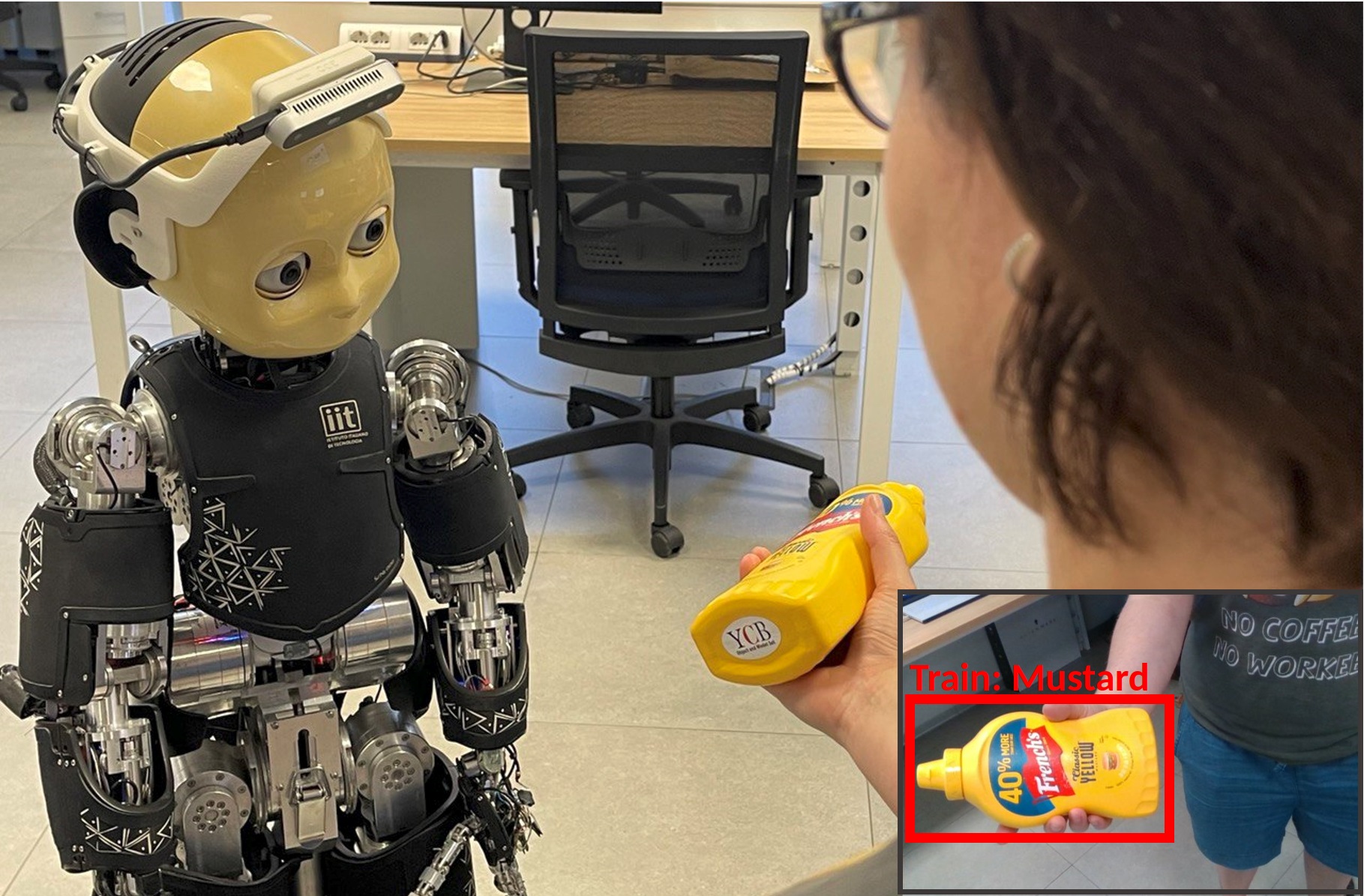}
    \caption{\textbf{Target human robot interaction setup.} The iCub is tracking the shown object by following the teacher's hand, while acquiring training samples automatically annotated with the proposed approach.}
    \label{fig:setup}
    \vspace{-0.5cm}
\end{figure}

Recent experimental developments suggest that perceiving social cues, and mainly direct eye-contact, seems to increase the TB effect~\cite{Ulloa2019,Vogel2021}. Eye contact is a fundamental social cue and plays a crucial role in non-verbal interaction between two or more individuals. The emergent literature on this topic supports the idea that eye-contact induces an enhancement of self-awareness paying more attention toward the self~\cite{Conty2016}. Gaze represents both a social signal to understand the readiness and the attention of the partner and an implicit action to influence the partner's behaviour, clearly linking the sense of agency to the mechanism of joint attention.
A crucial question is how the sense of agency may change in presence of embodied agents (and not only with other humans). Studies~\cite{Ciardo2020,Roselli2021} assess that interacting with social robotic agents affects SoA similarly to interacting with other humans, but differently from interacting with non-social mechanical devices (e.g. air pump).

Along with the social component, endowing a robot with different social skills can lead to more unconstrained human robot interaction (HRI) pipelines in which, for instance, the robot can learn a new task by interacting with the human partner, acting as a teacher. So doing, robot's learning process may become more natural and robust to noisy environments. 
Specifically, in our work we enhance the iCub humanoid robot~\cite{icub} with different human social skills in order to teach it to detect novel objects in less controlled experimental scenarios. For example, face recognition is used to recognise the teacher among other people, the ability to detect mutual gaze is needed to understand the teacher's intention, whereas gaze following is useful to understand which object the teacher wants iCub to learn. To this aim, we refer to the on-line learning approach proposed in~\cite{Maiettini2019,Ceola2021} for object detection. Exploiting human interaction~\cite{Maiettini2017,Maiettini2020}, such a  method is able to train or adapt an object detection model on-line avoiding tedious waits and allowing a faster and more reactive learning process. In this work, we improve the pipelines~\cite{Maiettini2017,Maiettini2020} by relaxing some crucial constraints during the interaction, making it more natural.

Furthermore, for a successful interaction the social cues have to be accurately timed. Studies revealed that in a non-verbal context direct eye-contact should last approximately $3-4$ seconds at a time. Any longer time may cause the other person feel uncomfortable~\cite{Helminen2011,Moyer2016}. In such studies several participants were asked to indicate their comfort level while watching different face stimuli who appeared to be looking directly at them for periods of time of variable length.
Argyle and Dean proposed that there is an equilibrium point between the duration of the eye-contact and the physical proximity between the coactors. They assessed that the duration of the eye-contact increases with distance. Thus, in average for a glance of $5$ seconds long, the distance should be $60$cm~\cite{Argyle1965}. During an interaction, the duration of the eye-contact even decreases. For example, according to~\cite{Argyle2017}, when two people are having a conversation on an emotionally neutral topic the length of eye contact is $1.5$s on average.

Summarising, in our work we aim at combining the dynamics and the patterns of the social interaction with the technicalities of the object detection algorithm to reach an effective on-line collaborative learning process between iCub and the human teacher. The rest of the paper is organised as follows. In Sec.~\ref{sec:related_work}, we overview the state of art on robots having social skills and on HRI-based learning pipeline. The proposed architecture based on social learning strategy is described in Sec.~\ref{sec:architecture}. Our empirical analysis is shown in Sec.~\ref{sec:training} and~\ref{sec:results}. Finally conclusions are drawn in Sec.~\ref{sec:conclusions}.

\begin{figure*}[thpb]
    \centering
    \includegraphics[trim={0 2cm 0 1.5cm}, clip, width=0.95\textwidth]{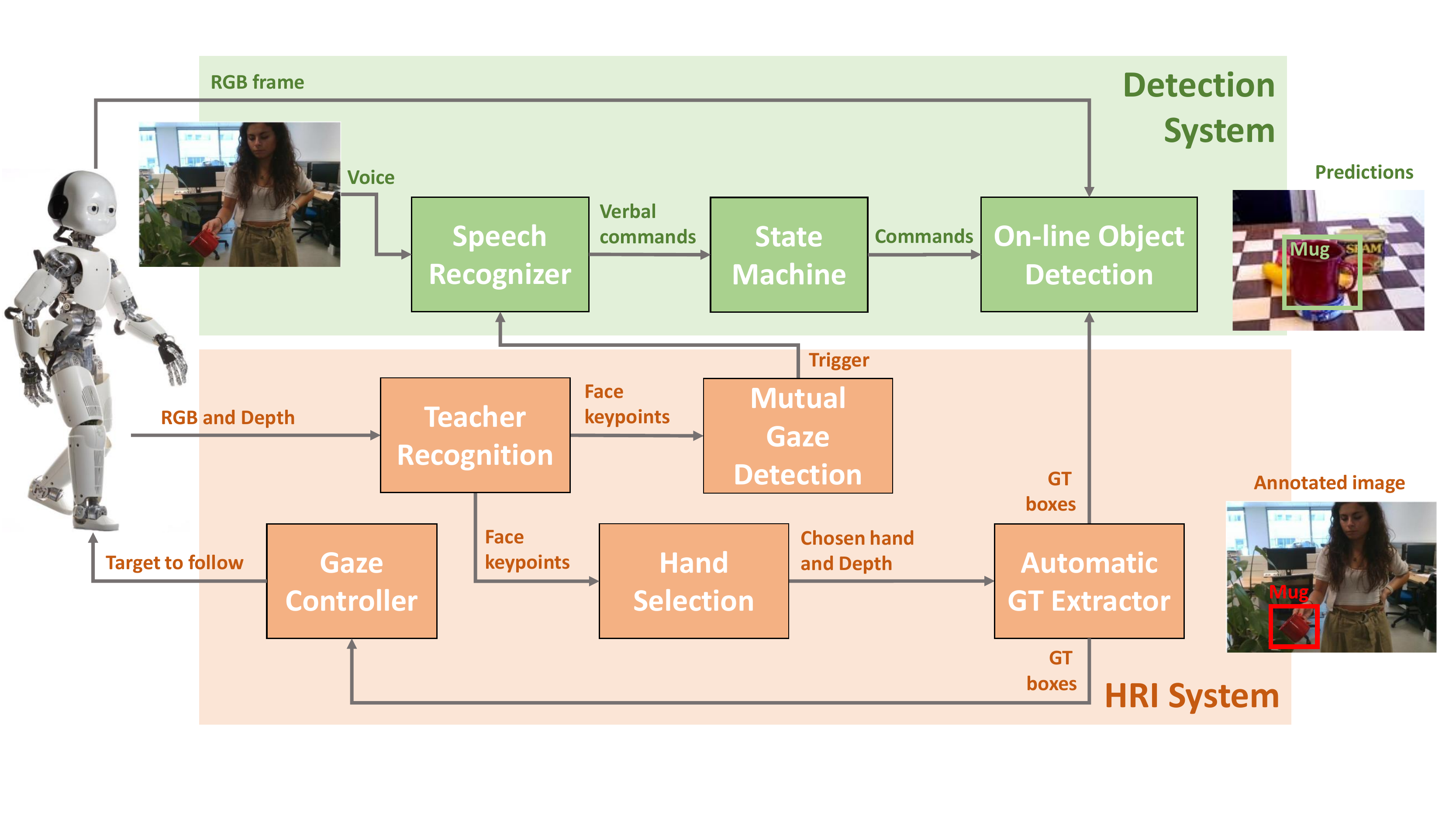}
    \caption{\textbf{Architecture of the proposed pipeline.} It is composed of two sub-systems: the green block represents the \textit{Detection System} and the red block represents the \textit{HRI System}. Our contribution relies on the latter one, endowing iCub with social skills such as face recognition, mutual gaze detection and gaze estimation that allow for a natural interaction with the teacher during the learning process. Refer to Sec.~\ref{sec:architecture} for further details.}
    \label{fig:architecture}
    \vspace{-0.5cm}
\end{figure*}

\section{RELATED WORKS}
\label{sec:related_work}
\subsection{Robots as social agents}
Enabling robots to exhibit social skills could lead to smoother and more effective HRI where humans and robots actually interact as peers \cite{Hancock2011,Thepsoonthorn2018}. For example, a social robot can help a human companion in several daily activities in a better manner by predicting their intention simply observing where their attention is focused at. Humans, indeed, tend to look at an object before trying to grasp it with the hand~\cite{Voudouris2018}. Gaze estimation plays a crucial role as social cue in many cooperation tasks~\cite{Kleinke1986,Emery2000}.

Studies in social robotics identified several main characteristics that users described as factors for a robot to appear and be accepted as social~\cite{deGraaf2015,Henschel2021}. Among them, interesting factors were the capability (1) of reacting to a human's stimuli (e.g. gesture, pointing, gaze); (2) of showing emotions and feelings; (3) of being autonomous and (4) of being socially aware of the environment. Several studies exist in the current literature addressing the aforementioned points. For example,~\cite{Rincon2019} proposed the social robot \textit{EmIR} which was developed to provide assistance to elderly people presenting human-like features like perceiving and displaying human emotions. Specifically, they trained a convolutional neural network to detect faces on the input image and classify them according to seven classes representing specific human emotions (i.e., fear, angry, upset, happy, neutral, sad and surprised). A lot of effort has been spent in implementing socially aware robotic behaviour in terms of navigation (refer to~\cite{Rios-Martinez2015} for a review). Among them, an extended version of the social force model for reactive navigation that integrates additional social cues used by humans while walking was proposed in~\cite{Reddy2021}. Specifically, the motion planning part of the proposed approach took into account some general social conventions, like for instance, that humans prefer to avoid each other from the left or they allow aggressive people to overtake by not coming on their way.
In another work~\cite{Tolgyessy2017}, the human can command an autonomous robot to navigate to a desired destination by raising one hand in order to trigger the robot and point to the final location with the other hand.

Humanoid robots with social abilities can be useful in various contexts. They can also provide support in the Healthcare, improving the quality of life of many people suffering of both motor impairments (e.g., after having a stroke)~\cite{Assad-Uz-Zaman2019,Mohebbi2020,Rosati2010} and social disorders like autism~\cite{Billard2007,Saleh2021}. Moreover, they can be used as research tool for a better understanding of ourselves as humans~\cite{Wykowska2020,Wykowska2021} exploring social concepts like trust~\cite{Stower2021}, empathy~\cite{García-Corretjer2022,Vinanzi2021} and attachment~\cite{Dziergwa2018}.

\subsection{HRI for vision task learning}
Current deep learning based methods proved to be effective for obtaining general purpose object detection models. However, gathering the sufficient ground truth for training them through supervised learning is a costly operation. It requires drawing a bounding box around each object of interest, providing the corresponding label, in each image of the training set (which, depending on the task, might be typically composed of thousands of them). One possibility to overcome this issue is to exploit the robot embodiment and its chance to interact with humans and the surrounding environment. In previous work, it has been shown that object detection methods (and specifically the one proposed in~\cite{Maiettini2019}) can be trained with human interaction~\cite{Maiettini2017,Maiettini2020} in a constrained setting. Specifically, the human shows the novel object to be trained to the robot and the information from the robot's depth sensors is used to localise the object and follow it with the robot’s gaze. The latter can be segmented and the corresponding bounding box automatically assigned and gathered as ground truth. The main assumption of this approach is that the shown object is the closest element to the robot's camera. In fact, iCub takes the closest blob of pixels regardless it is the object effectively handled by the teacher. Such an issue can have a strong impact in crowded environments where other people can stand near the robot (for instance, in working places). This forces the teacher to be very close to the camera and to keep a specific behaviour during the training process, preventing a na\"ive user to perform it. In this work, we consider it as a starting point for our work. We endow the iCub humanoid with social skills (such as face recognition, mutual gaze detection and gaze estimation) with the aim to relax the interaction constraints considered in~\cite{Maiettini2017,Maiettini2020} and allow for a more natural interaction in the teacher-learner setting. We show with experiments that our approach make the data acquisition pipeline more robust to common causes of noise of the surrounding environment.



\section{PIPELINE WITH SOCIAL CUES}
\label{sec:architecture}
The proposed approach allows to exploit social cues to reduce the assumptions on the interaction required in the teacher-learner setting (Fig.~\ref{fig:setup} represents the considered setup). Specifically, based on social studies~\cite{Conty2016, Argyle2017, Voudouris2018}, we devise the learning pipeline as follows. The iCub recognises the teacher among all presents and reacts to their prolonged eye-contact ($\sim1.5$s) by engaging the interaction and waiting for verbal commands. At this point, the teacher instructs iCub on which object to learn by pronouncing its name and by showing it in their hand while looking at it. The robot, following the teacher gaze, identifies the object and starts tracking it acquiring labelled frames. Finally, iCub uses the collected data to train an object detection model.
Note that, the proposed approach removes all assumptions on the shown object position with respect to the camera. The teacher can be at any distance and other elements or persons can be between the camera and the shown object. Moreover, the teacher gaze interpretation allows to eliminate any explicit trigger (e.g., microphone un-mute, hand selection) for starting parsing the user commands or identifying the hand holding the object. This permits to achieve a natural HRI, improving the downstream object detection learning.

\subsection{Pipeline overview}
The proposed architecture (depicted in Fig.~\ref{fig:architecture}) is composed of a \textit{Detection system} and an \textit{HRI system}. The former one implements the on-line object detection method and training pipeline of~\cite{Maiettini2019,Maiettini2020}, while the latter integrates social skills (namely, face recognition, mutual gaze detection and gaze estimation) in the architecture.

The \textit{Teacher recognition} is the first module of the proposed HRI pipeline. It endows iCub with the ability to recognise the teacher among all the people in the scene. It receives the RGB and the depth images from the robot's camera and outputs the body key-points belonging to the teacher. This module allows iCub to focus his attention on the teacher even in an environment with multiple people. The teacher's key-points are taken as input by the \textit{Mutual gaze detection} module which recognises when the teacher wants to start an interaction with the robot by establishing eye-contact with it. Once iCub has detected the will of the teacher of starting an interaction, it focuses its attention on the teacher, waiting verbal commands. Thanks to the \textit{Speech recogniser} module, the teacher instructs iCub on which object it is required to learn simply by speaking. The teacher body/face key-points extracted by the \textit{Teacher recognition} are also used by the \textit{Hand selection} module to estimate the teacher gaze direction (left or right) that allows iCub to understand which is the teacher's hand holding the object of interest. The idea behind this module exploits the concept of joint attention during interaction. It assumes that a human tends to look at the objects they are speaking about. Such a social ability allows iCub to move its gaze toward the same object the human is looking at. Exploiting such a joint attention, the \textit{Automatic ground truth (GT) extractor} can select the blob of pixels belonging to the object held in the selected hand and start tracking it, associating it with the label given by the teacher through the verbal command. This blob of pixels, jointly with the label and the starting image, are used both as training examples by the \textit{On-line object detection} and for tracking the object with iCub's gaze during the interaction by the \textit{Gaze controller}. The entire behaviour is orchestrated by a \textit{State machine}.
All the aforementioned software modules with the sensors and actuators are connected through the open-source middleware YARP\footnote{\url{https://www.yarp.it}} (Yet Another Robot Platform)~\cite{Fitzpatrick2008}.
In the following paragraphs we describe how each of the main modules of the architecture was trained and built.

\begin{figure}[t]
    \centering
    \includegraphics[trim={7.6cm 5.1cm 8cm 3.5cm}, clip, width=\columnwidth]{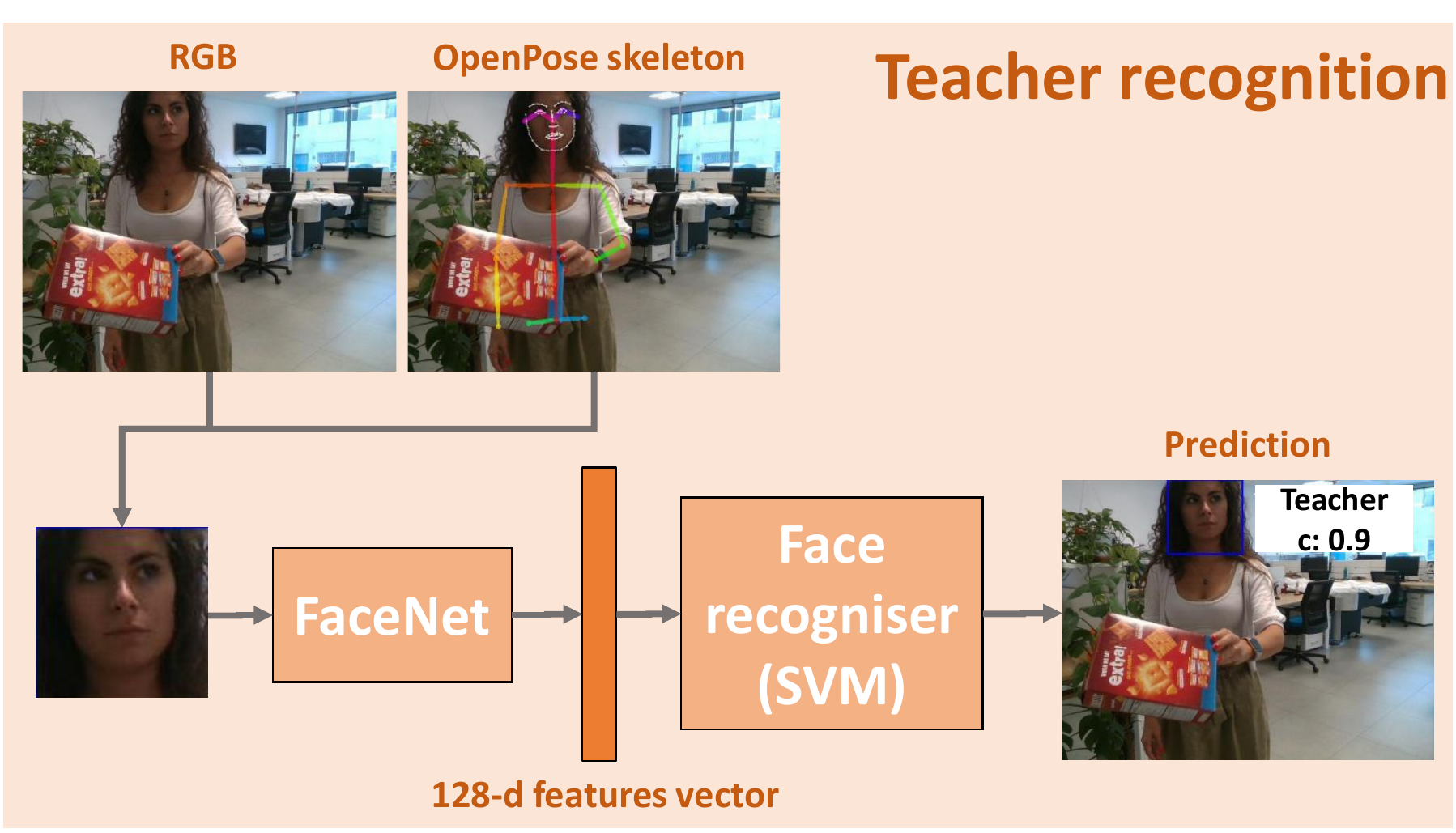}
    \caption{\textbf{Teacher recognition.} The image and OpenPose's output are used to localise and crop all faces. Convolutional embeddings are extracted and are used by the SVM classifier to recognise the teacher among all the presents.}
    \label{fig:face_recogniser}
    \vspace{-0.5cm}
\end{figure}

\subsection{Teacher recognition}
The \textit{Teacher recognition} (Fig.~\ref{fig:face_recogniser}) aims at recognising the teacher among all the people in the scene. Once it receives the RGB frame from the camera, OpenPose~\cite{Cao2019} is used to build the bounding box for each individual in the image. This is a well-known system for multi-human pose estimation. It predicts the 2D location, in pixel $(x,y)$, of the anatomical key-points of each person in the image with the corresponding confidence level $k$. Here, we extract only the face key-points in order to compute the bounding boxes and cut the input around the detected faces. The face thumbnails are given to the pre-trained network FaceNet~\cite{Schroff2015} to extract the corresponding face embeddings ($128$-D convolutional feature vector). FaceNet is a CNN with a triplet-based loss function proposed both for face recognition and clustering (we use it to extract the feature vector for each face). Then, the feature vectors are given to a binary Support Vector Machine (SVM) classifier with RBF kernel trained to recognise the teacher from other people in the scene. The hyperparameters of the SVM model were selected using a randomised search over a grid parameters. The module's output is the OpenPose skeleton that belongs to the teacher.

This module works completely online. It allows to acquire batches of sample data and train a new SVM model as the data increases (one batch has $300$ samples for each class). The samples (image crops of the face extracted with FaceNet as described above) are automatically labelled as \textit{teacher} whereas the samples belonging to the negative class are labelled randomly picking data from the LFW dataset\footnote{\url{http://vis-www.cs.umass.edu/lfw/}}. Such a dataset contains more than $6$k labelled faces and is the same used to train FaceNet.
The incremental batches of data acquired online are randomly split in train and validation set (the $30\%$ of the dataset is used as validation set).
The training ends when the accuracy on the validation set reaches a threshold of $0.99$. After that, the module uses the last saved model to produce its output.
The main advantage to have a face recogniser that can be trained online is that the system is not constrained to a specific subset of teachers pre-trained a-priori. Furthermore, to allow this module to efficiently work online in a real training scenario, a tracking system of the teacher's body has been implemented in case their face is not visible from the camera (for example, when iCub turns its head to track the object of interest). Specifically, we track the keypoint representing the teacher's hip comparing the relative distance between it and that of all the people in the scene in consecutive frames.

\begin{figure*}[thpb]
    \centering
    \includegraphics[trim={1.2cm 6.6cm 1.2cm 0cm}, clip, width=0.9\textwidth]{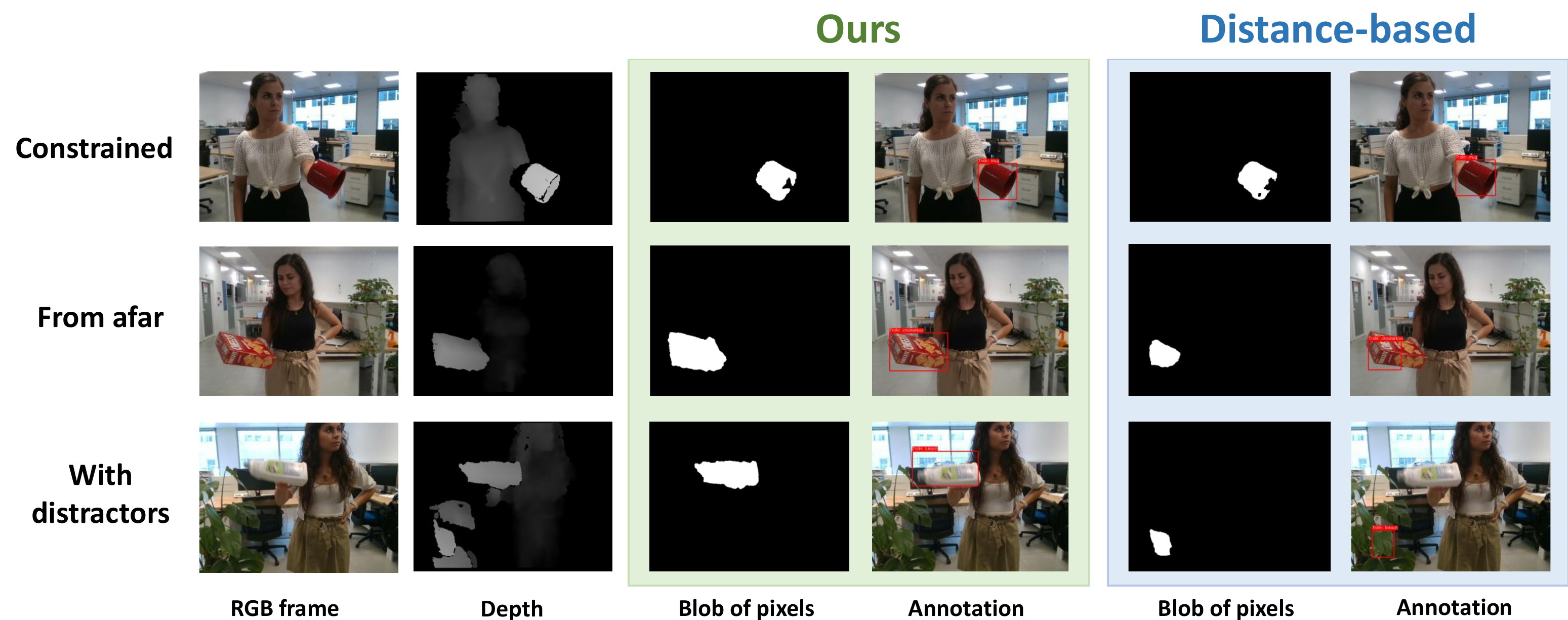}
    \caption{\textbf{Annotations examples.} We show automatically annotated example frames by \textit{Ours} and the \textit{Distance-based} system in the three scenarios. For each example, we show the RGB and depth images, the segmented pixels blob and the resulting computed annotation for both pipelines.}
    
    \label{fig:training}
        \vspace{-0.5cm}
\end{figure*}

\subsection{Mutual gaze detection}
For this module we rely on~\cite{Lombardi2022}. Briefly, such a classifier exploits OpenPose to extract the feature vector of the individual of interest (in the proposed application this is identified by the \textit{Teacher recognition} module). A subset of $19$ face key-points are considered  ($8$ points for each eye, $2$ points for the ears and $1$ for the nose) resulting in a feature vector of $57$ elements (i.e. the triplet $(x,y,k)$ is taken for each point). Then, a binary SVM classifier is trained to recognise events of \textit{eye-contact / no eye-contact}. For details, refer to~\cite{Lombardi2022}.

\subsection{Hand selection}
The \textit{Hand selection} module was trained using the RGB frames collected in~\cite{Lombardi2022} with the iCub cameras. This dataset is composed of images from $24$ participants that move their gaze in different directions. In this work, we manually labelled with \textit{left / right} each frame and we used them to train a gaze direction estimation model, that discriminating the teacher gaze direction, allows the robot to focus on the hand a human is looking at.

Each RGB frame acquired from the camera is given as input to OpenPose in order to extract the feature vector. As for the \textit{Mutual gaze detection}, we consider the $19$ teacher face key-points as feature vector. This feature vector is centred with the respect to the head centroid and normalised on the farthest point from it.
The resulting feature vector is finally used to train a two-classes SVM with RBF kernel. Specifically, the hyperparameters of the SVM model are selected using a 5-fold grid search cross-validation. 
For the training of the classifier, we split the dataset such that $19$ out of the total $24$ participants are used as training set, while the remaining $5$ as test set for performance evaluation. We repeat the split for $5$ trials in order to average the performance over different subsets of participants. The reported performances over the test sets are: accuracy $= 0.99 \pm 0.01$, precision $= 0.98 \pm 0.02$, recall $= 1.00 $, F1-score $= 0.99 \pm 0.01$.
At inference time, the module's output is the pair $(p,c)$ where $p$ is the prediction of the hand the human is looking at, while $c \in [0,1]$ is the confidence level.

\subsection{On-line object detection}
For this module we rely on the method presented in~\cite{Maiettini2019}. Given an input image, it consists of (i) a first stage of region proposals and feature extraction and (ii) a second stage of region classification and bounding box refinement. The former relies on layers from Mask R-CNN~\cite{He2017}. They are used to extract a set of Regions of Interest (RoIs) from the input image and encode them into convolutional features. In this work, we considered ResNet50~\cite{He2016} as the convolutional backbone for Mask R-CNN. The second part classifies and refines the RoIs proposed at the previous stage by using respectively a set of FALKON~\cite{Rudi2017} binary classifiers (one for each object class to detect) for the classification and Regularized Least Squares (RLS) for the refinement. Specifically, the training of the classifiers applies an approximated bootstrapping approach, called Minibootstrap~\cite{Maiettini2019} to overcome the background-foreground class imbalance problem in object detection, keeping a fast learning. Refer to~\cite{Maiettini2019} for details.

\subsection{Automatic GT extractor}
The main drawback of previous work on HRI systems for robot learning, like e.g.~\cite{Maiettini2017,Maiettini2020}, is the list of assumptions regarding the human teacher. In~\cite{Maiettini2020}, for instance, in the teacher-learner setting, the main hypothesis is that the shown object is the closest to the robot's camera. In fact, a tracking routine~\cite{Pasquale2016} selects the pixels from the depth map that are closer to the robot, segmenting them from the background. However, this hypothesis limits the usage range to less than $1$m. The reason for this is twofold: firstly, considering relatively small objects, the blob of pixels depicting them from a distance that is further than $1$m is so small that the noise from the depth sensor might generate problems in the localisation of the closest pixels; secondly, with a greater distance between the robot and the teacher it is more difficult to satisfy the requirement of having the object as the closest element to the robot's camera. 
For these reasons, in this work we relax this assumption, by exploiting information about the human teacher. Specifically, the positions of the teacher body’s key-points (given by considering the output of OpenPose jointly with that of the \textit{Teacher recognition} module) are used to drive robot’s attention toward the teacher. Precisely, we exploit the fact that the shown object’s 3D points are close to the teacher’s hand. Therefore, knowing the precise position of this latter allows to efficiently segment the held object using depth information.

\section{EXPERIMENTAL SETUP}
\label{sec:training}
With the aim of showing the effectiveness of the proposed approach with respect to the one presented in~\cite{Maiettini2017}, we compare them in three different scenarios, with different levels of noise in the environment. In the rest of the paper, we refer our architecture as \textit{Ours} whereas the one proposed in~\cite{Maiettini2017} as \textit{Distance-based system}. The three different scenarios are:
\begin{enumerate}
    \item \textbf{Constrained}: it is the simplest scenario. Only the teacher is in the scene at less than $1$m far from the robot ($\sim0.5/0.8$m). Moreover, the object to learn is always the closest element to the robot's camera;
    \item \textbf{From afar}: this scenario represents the use case where the object is shown at a grater distance from the robot's camera ($\sim2$m). So doing, the object (which is still always the closest element to the robot) appears smaller and with a lower amount of detail.
    \item \textbf{With distractors}: in this last scenario, the teacher is $\sim 2$m far from the robot, as in the previous one, and is not alone in the scene. There are other objects and persons (called distractors) that can appear behind the camera or between the camera and the shown object. This scenario represents a more natural, less constrained, training setting.
\end{enumerate}
For our experiments, we considered $9$ objects from the YCB dataset~\cite{Calli2015} (namely, \textit{011\_banana, 021\_bleach\_cleanser, 003\_cracker\_box, 035\_power\_drill, 025\_mug, 006\_mustard\_bottle, 010\_potted\_meat\_can, 037\_scissors, 004\_sugar\_box}). To benchmark the system, we show each object to the robot, one at a time, in all the aforementioned scenarios, acquiring three different sets of image sequences (one for each scenario). Every set is composed of $9$ sequences (one for each considered object) having $300$ frames each (acquired at $\sim7$ frames per second). We use these sets of images to compare performance of \textit{Ours} and  \textit{Distance-based system}. Firstly, we use them to evaluate the quality of the annotations provided by both approaches (Sec.~\ref{sec:experiments:annotation}). Then, we use these annotations, together with the images, to train the \textit{On-line object detection} module. We compare object detection performance of models trained with ground truth from \textit{Ours} and the \textit{Distance-based system} (Sec.~\ref{sec:experiments:clean} and~\ref{sec:experiments:noise}). To this aim, we test the models in a twofold way. Formerly, we show again the $9$ objects to iCub in the same conditions considered for the \textbf{Constrained} set (\textit{iCub test set}) and we evaluate the quality of the object detection performance on $900$ frames ($100$ for each object) by comparing the predicted detections with a manual ground truth (note that, manual annotation of the sequences of the \textit{iCub test set} was required to obtain ground-truth for benchmarking purposes, i.e. the ground-truth acquired in this way was not used for training). Lastly, as a further validation, we test the models trained with \textit{Ours} and the \textit{Distance-based system} on a subset of the Ho3D dataset~\footnote{\url{https://github.com/shreyashampali/ho3d}}~\cite{Hampali2020}. For doing that, we randomly selected one sequence for each of the $9$ objects (\textit{ABF11}, \textit{BB14}, \textit{GPMF14}, \textit{GSF14}, \textit{MC1}, \textit{MDF11}, \textit{SiS1}, \textit{SM2}, \textit{SMu1}).

We report performance in terms of mAP (mean Average Precision) with the IoU (Intersection over Union) threshold
set to $0.5$, as defined for COCO~\cite{COCO}, using the publicly available implementation~\footnote{\url{https://github.com/cocodataset/cocoapi/tree/master/PythonAPI/pycocotools}}.


\section{RESULTS}
\label{sec:results}
While the video submitted as supplementary material shows the functioning of the proposed application, in this section, we demonstrate its effectiveness reporting on the performed empirical evaluation.

\subsection{Annotations quality evaluation}
\label{sec:experiments:annotation}
Firstly, we aim at assessing the quality of the automatic annotations produced by the proposed approach. We do that by randomly sub-sampling $\sim270$ frames ($\sim30$ for each object) from each of the three sets of sequences described in the previous section (namely, \textit{Constrained}, \textit{From afar}, \textit{With Distractors}) and we manually annotated them. We compare this manual ground truth with the automatic annotations produced respectively by \textit{Ours} and the \textit{Distance-based system}, to asses their quality (refer to Fig.~\ref{fig:training} for annotation examples from both approaches). Tab.~\ref{table:annotations} reports the obtained results.
As it can be noticed, in the \textbf{Constrained} condition, our approach outperforms the baseline of $\sim11\%$. Such a gap is mainly given by the different precision of the two systems for big objects, like for instance, the\textit{ 003\_cracker\_box} and the \textit{035\_power\_drill} (e.g., for the  \textit{003\_cracker\_box} our approach and the baseline achieve respectively an mAP of $92.6$\% and $70.7$\%). Indeed, depending on their pose with respect to the camera, those objects can be represented by a cluster of points in the image with significantly different distances from the depth sensor. In these cases, knowing the distance of the teacher's hand from the camera, is crucial to identify the image area to focus on and correctly segment all the pixels belonging to the held object. On the contrary, sharply thresholding on the depth level, without any other contextual knowledge (which is done in the  \textit{Distance-based system}) might lead to an imprecise segmentation. This might be addressed, to some extent, with an ad hoc hyper-parameters tuning of the \textit{Distance-based system}, based on the size of the considered object and its distance from the camera. Nevertheless, a more general method that does not need any per-object adaptation, is preferable.
Moreover, for the two more challenging conditions where the objects are shown from a greater distance and with distracting elements (namely, \textbf{From afar} and \textbf{With distractors}), the performance obtained by the proposed approach is significantly higher than the baseline. This shows the proposed method effectiveness in reducing the impact of disturbance elements during the data acquisition.
Finally, these results demonstrate that the proposed method is able to automatically annotate the desired object in the image with a high mAP. Note that, even if this is not $100\%$, the obtained annotations can be used to train an object detection model. Indeed, it has been shown in previous work~\cite{Maiettini2017,Maiettini2020} that learning based object detection models are able to average out a certain level of noise in the dataset annotations.
In the next sections, we evaluate performance of the entire proposed HRI-based learning pipeline. 

\begin{table}[t]
	\centering
	\begin{tabular}{c|c|c|c}
		\cline{1-4}
		  & \textbf{\shortstack{Constrained \\ mAP (\%)}}    & \textbf{\shortstack{From afar \\ mAP (\%)}} & \textbf{\shortstack{With distractors \\ mAP (\%)}}  \\ \hline
	  	  \textbf{Distance-based}    & 73.0   &  51.2 & 19.3  \\ \hline
		  \textbf{Ours}      &  84.4    & 70.5  &  68.0 \\ \hline
	\end{tabular}
	\caption[]{Annotations quality evaluation. We compare annotations produced by Ours and Distance-based system.}
	\label{table:annotations}
    \vspace{-0.8cm}
\end{table}

\subsection{Object detection in a constrained scenario}
\label{sec:experiments:clean}
In this section, we analyse the object detection performance of models trained with the proposed approach. We compare them with models trained with the \textit{Distance-based system}, reported as reference baseline. The aim here is to verify the effectiveness of the proposed HRI-based learning pipeline when all the assumptions required by the baseline hold (i.e., close distance from the robot and no disturbance elements). 
To this aim in our experiments, we use the \textbf{Constrained} set of image sequences. We split the objects of the dataset by their size, identifying three different sizes: \textit{Small} (\textit{025\_mug, 011\_banana, 010\_potted\_meat\_can}), \textit{Medium} (\textit{004\_sugar\_box, 006\_mustard\_bottle, 037\_scissors}) and \textit{Big} (\textit{003\_cracker\_box, 021\_bleach\_cleanser, 035\_power\_drill}) and we train three different models one for each objects size. We do that to evaluate the impact of objects size in the automatic annotations collection and in the subsequent object detection training. We test the obtained model on the \textit{iCub test set} (see Sec.~\ref{fig:training}).
Results are  reported in Tab.~\ref{table:test_set_clean}. As it can be noted, for small and medium sized objects the models trained with the two different pipelines perform almost equivalently. This is due to the fact that the \textbf{Constrained} scenario satisfies all hypotheses required by the \textit{Distance-based system} to function well (i.e., close distance from the robot and no disturbance elements). However, for bigger objects, the model trained with \textit{Ours} performs significantly better. This is caused by the fact that, as explained in the previous section, the \textit{Distance-based system} has issues in segmenting big objects due to the lack of contextual information. This annotation issue produces the object detection performance gap reported in the third column of Tab.~\ref{table:test_set_clean} for big objects.

\begin{table}[t]
	\centering
	\begin{tabular}{c|c|c|c}
		\cline{1-4}
		  \textbf{\shortstack{Objects size\\ { }}  }          & \textbf{\shortstack{Small\\ mAP (\%)}} & \textbf{\shortstack{Medium\\ mAP (\%)}} & \textbf{\shortstack{Big\\ mAP (\%)}}   \\ \hline
	  	  \textbf{Distance-based}    & 73.2  & 63.4     & 58.5    \\ \hline
		  \textbf{Ours}      & 72.4    & 65.4     & 80.7    \\ \hline
	\end{tabular}
	\caption[]{Performance on the iCub test set, of models trained on the \textbf{Constrained} dataset. We compare models trained with Ours and Distance-based system.}
	\label{table:test_set_clean}
    \vspace{-0.8cm}
\end{table}

\begin{table}[]
	\centering
 	\vspace{0.25cm}
	\resizebox{8.4cm}{!}{
    	\begin{tabular}{c|c|c|c|c|c|c}
    		\cline{1-7}
    		                                                   & \multicolumn{3}{c|}{\textbf{\shortstack{From afar}}} & \multicolumn{3}{c}{\textbf{\shortstack{With distractors}}}  \\ \hline
    		  \textbf{\shortstack{Object size\\ { } }}                &  \textbf{\shortstack{Small\\ mAP (\%)}} & \textbf{\shortstack{Medium\\ mAP (\%)}} & \textbf{\shortstack{Big\\ mAP (\%)}}   &  \textbf{\shortstack{Small\\ mAP (\%)}} & \textbf{\shortstack{Medium\\ mAP (\%)}} & \textbf{\shortstack{Big\\ mAP (\%)}}  \\ \hline
    	  	  \textbf{\shortstack{Distance-based}}    &  43.0    & 55.3     & 24.0  &   29.0     &   37.0     & 32.0  \\ \hline
    		  \textbf{\shortstack{Ours}}              &  47.3    & 63.9     & 74.6  &   52.9     &    59.4    & 66.8 \\ \hline
    	\end{tabular}
	    }
	\caption[]{Performance on iCub test set, of models trained on \textbf{From afar} and \textbf{With distractors} datasets. We compare models trained with Ours and Distance-based system.}
	\label{table:test_set}
	\vspace{-0.8cm}
\end{table}
 
\subsection{Object detection in a noisy environment}
\label{sec:experiments:noise}
Finally, in this section, we remove the assumptions required by the \textit{Distance-based system} and we prove that in these conditions, the proposed approach is critical to obtain good object detection performance.
For doing this, we consider the \textbf{From afar} and \textbf{With distractors} sets of image sequences, we split them according to the objects sizes as described in the previous section and we use the resulting datasets for training object detection models. 
We test the obtained models on the \textit{iCub test set} and, as a further validation, on a subset of the Ho3D dataset (Sec.~\ref{sec:training}).

Results are reported in Tab.~\ref{table:test_set} for the \textit{iCub test set} and in Tab.~\ref{table:hord} for Ho3D. As it can be noticed, in general the models trained with the proposed approach perform better than the baseline on both test sets.
Note that, in both cases the performance for small objects is the lowest. This is due to the fact that in the two test sets all the objects are depicted at a close distance, while in the two considered training sets they appear from a afar. This main difference especially affects performance for small objects because their appearance in the image is the most altered when taken from afar. Moreover, one may note that the average object detection performance in both scenarios is lower than the one obtained by using the \textbf{Constrained} training set. This is due to the fact that the \textit{iCub test set} has been acquired in the same conditions used for the collection of \textbf{Constrained}, while the other two sets (\textbf{From afar} and \textbf{With distractors}) have been acquired in different conditions. This produces the so called domain shift~\cite{Maiettini2019b} between the train and the test sets, resulting in the reported performance drop.
Therefore, this is not caused by the annotations quality and it has been shown in previous work~\cite{Maiettini2020_thesis,Maiettini2019b,Maiettini2020} that it can be solved by, for instance, integrating (i) a robot's autonomous exploration of the new domain and (ii) weakly-supervised learning techniques.

\begin{table}[]
	\centering
	\resizebox{8.4cm}{!}{
    	\begin{tabular}{c|c|c|c|c|c|c}
    		\cline{1-7}
    		                                    & \multicolumn{3}{c|}{\textbf{\shortstack{From afar}}} & \multicolumn{3}{c}{\textbf{\shortstack{With distractors}}}  \\ \hline
    		   \textbf{\shortstack{Object size\\ { }}}               &  \textbf{\shortstack{Small\\ mAP (\%)}} & \textbf{\shortstack{Medium\\ mAP (\%)}} & \textbf{\shortstack{Big\\ mAP (\%)}}  &  \textbf{\shortstack{Small\\ mAP (\%)}} & \textbf{\shortstack{Medium\\ mAP (\%)}} & \textbf{\shortstack{Big\\ mAP (\%)}}  \\ \hline
    	  	  \textbf{\shortstack{Distance-based}}    & 30.8 & 58.3 & 25.0 & 26.3 & 47.0 & 25.0 \\ \hline
    		  \textbf{\shortstack{Ours}}              & 42.5 & 72.2 & 74.8 & 58.2 & 56.4 & 43.9 \\ \hline
    	\end{tabular}
	}
	\caption[]{Performance on Ho3D of models trained on the \textbf{From afar} and \textbf{With distractors} datasets. We compare models trained with Ours and Distance-based system.}
	\label{table:hord}
    \vspace{-0.8cm}
\end{table}

\begin{figure}[h!]
\vspace{0.25cm}
    \centering
    \includegraphics[width=0.95\columnwidth]{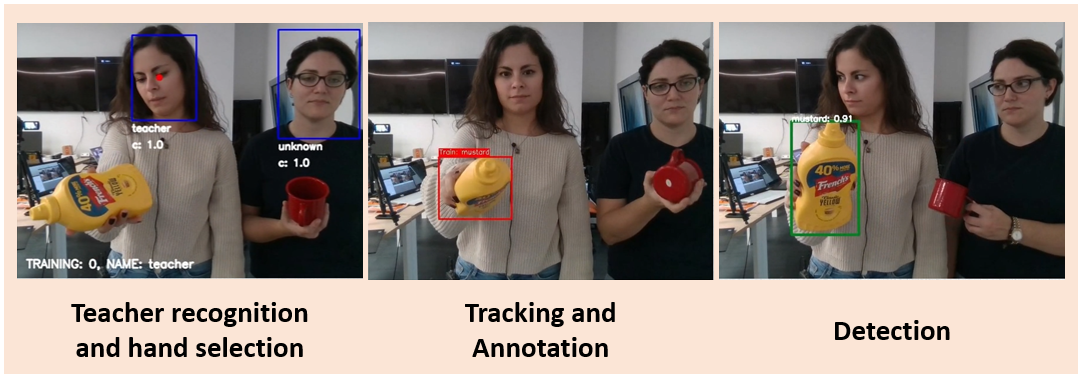}
    \caption{\textbf{Annotation with multiple humans.} Sample video frames representing the teacher recognition and the annotation process in presence of two different humans in the scene.}
    \label{fig:video}
    \vspace{-0.5cm}
\end{figure}

In conclusion, the proposed approach proved to be effective for object detection training, being critical especially when the data acquisition conditions are not ideal. This allows for a more natural interaction with the teacher and to be more robust to distraction elements than the baseline.

\subsection{Object detection with multiple humans}
For the sake of brevity we addressed the scenario with different humans in the video attached to the paper. We recorded the whole training pipeline having two people in the iCub's field of view proving that iCub is able to recognise the teacher and segment only the object in selected hand. Here we depict some sample frames in the Fig.\ref{fig:video}.

\section{CONCLUSIONS}
\label{sec:conclusions}
In this paper, we propose to exploit social cues to improve current learning pipelines on robotic platforms. We present an architecture to train iCub to detect new objects in a teacher-learner setting, exploiting different human social cues. We focus on the social cues of the mutual gaze, gaze following and teacher recognition, overcoming several issues arose in previous work (mostly due to constrains in the interaction with the human). Specifically, we compare our approach with the learning pipeline in~\cite{Maiettini2017,Maiettini2020} reporting an improvement in performance in different environmental conditions. By removing several interaction constraints, our approach allows for a more natural human interaction with the robot. Therefore, our architecture can be used in less controlled setups, also in the experimental psychology, and might open new directions to study when a human feels engaged with a robotic partner. For example, in the setup of this work we can manipulate the delay in the actions/reactions of iCub to shed the light on the joint sense of agency.



\section*{Acknowledgment}
This work was supported in part by the Istituto Nazionale Assicurazione Infortuni sul Lavoro, under the project iHannes (PR19-PAS-P1).


\begin{thebibliography}{99}

\bibitem{Gallagher2000} S. Gallagher, Philosophical conceptions of the self: implications for cognitive science., in Trends in cognitive sciences, 4(1), 2000.
\bibitem{Haggard2017} P. Haggard, Sense of agency in the human brain, in Nature Reviews Neuroscience, 18(4), 2017, pp.196-207.
\bibitem{Barlas2013} Z. Barlas, and S. Obhi, Freedom, choice, and the sense of agency, in Frontiers in human neuroscience, 7, 2013, p. e514.
\bibitem{Ulloa2019} J.L. Ulloa, R. Vastano, N. George, and M. Brass, The impact of eye contact on the sense of agency, in Consciousness and cognition, 74, 2019, p. e102794.
\bibitem{Vogel2021} D.H. Vogel, M. Jording, C. Esser, P.H. Weiss, and K. Vogeley, Temporal binding is enhanced in social contexts, in Psychonomic Bulletin \& Review, 28(5), 2021, pp. 1545-1555.
\bibitem{Conty2016} L. Conty, N. George, and J.K. Hietanen, Watching Eyes effects: When others meet the self, in Consciousness and cognition, 45, 2016
\bibitem{Ciardo2020} F. Ciardo, F. Beyer, D. De Tommaso, and A. Wykowska, Attribution of intentional agency towards robots reduces one’s own sense of agency, in Cognition, 194, 2020, p. e104109.
\bibitem{Roselli2021}  C. Roselli, F. Ciardo, and A. Wykowska, Intentions with actions: The role of intentionality attribution on the vicarious sense of agency in Human–Robot interaction, in Quarterly Journal of Experimental Psychology, 75(4), 2022, pp. 616–632.
\bibitem{icub} G. Metta, L. Natale, F. Nori, G. Sandini, D. Vernon, L. Fadiga, C. Hofsten, K. Rosander, M. Lopes, J. Santos-Victor, A. Bernardino, and L. Montesano, The iCub humanoid robot: an open-systems platform for research in cognitive development, in Neural networks : the official journal of the International Neural Network Society, 23(8-9), 2010.
\bibitem{Maiettini2019} E. Maiettini, G. Pasquale, L. Rosasco, and L. Natale, On-line object detection: a robotics challenge, in Autonomous Robots, 2019.
\bibitem{Ceola2021} F. Ceola, E. Maiettini, G. Pasquale, L. Rosasco, and L. Natale, Fast object segmentation learning with kernel-based methods for robotics, in IEEE International Conference on Robotics and Automation, 2021.
\bibitem{Maiettini2017} E. Maiettini, G. Pasquale, L. Rosasco, and L. Natale, Interactive data collection for deep learning object detectors on humanoid robots, IEEE-RAS 17th International Conference on Humanoid Robots, 2017.
\bibitem{Maiettini2020} E. Maiettini, V. Tikhanoff, and L. Natale, Weakly-Supervised Object Detection Learning through Human-Robot Interaction, in IEEE-RAS 20th International Conference on Humanoid Robots, 2020.
\bibitem{Helminen2011} T.M. Helminen, S.M. Kaasinen, and J.K. Hietanen, Eye contact and arousal: the effects of stimulus duration, in Biological Psychology, 88, 2011, p. 124–130.
\bibitem{Moyer2016} M.W. Moyer, Eye Contact: How Long Is Too Long? in Scientific American Mind, 27(1), 2016, p. e8.
\bibitem{Argyle1965} M. Argyle, and J. Dean, Eye contact, distance and affiliation, in Sociometry, 28, 1965, pp. 289-304.
\bibitem{Argyle2017} M. Argyle, Bodily communication, 2nd edition, Routledge editor, 2017.
\bibitem{Hancock2011} P.A. Hancock, D.R. Billings, K.E. Schaefer, J.Y. Chen, E.J. De Visser, and R. Parasuraman, A meta-analysis of factors affecting trust in human-robot interaction, in Human factors, 53(5), 2011, pp.517-527.
\bibitem{Thepsoonthorn2018} C. Thepsoonthorn, K.I. Ogawa, and Y. Miyake, The relationship between robot’s nonverbal behaviour and human’s likability based on human’s personality, in Scientific reports, 8(1), 2018, pp. 1-11.
\bibitem{Voudouris2018} D. Voudouris, J.B. Smeets, K. Fiehler, and E. Brenner, Gaze when reaching to grasp a glass, in Journal of vision, 18, 2018, p. e16
\bibitem{Kleinke1986} C.L. Kleinke, Gaze and eye contact: a research review, in Psychological bulletin, 100, 1986, p. e78
\bibitem{Emery2000} N.J.Emery, The eyes have it: the neuroethology, function and evolution of social gaze, in Neuroscience \& biobehavioral reviews, 24, 2000.
\bibitem{deGraaf2015} M.M. de Graaf, S. Ben Allouch, and J.A. van Dijk, What makes robots social?: a user’s perspective on characteristics for social human-robot interaction, in International Conference of Social Robotics, Springer , 2015, pp. 184–193.
\bibitem{Henschel2021} A. Henschel, G. Laban, and E.S. Cross, What makes a robot social? a review of social robots from science fiction to a home or hospital near you, in Current Robotics Reports, 2(1), 2021, pp. 9-19.
\bibitem{Assad-Uz-Zaman2019} M. Assad-Uz-Zaman, M. Rasedul Islam, S. Miah, and M.H. Rahman, NAO robot for cooperative rehabilitation training, in Journal of Rehabilitation and Assistive Technologies Engineering, 6, 2019, p. e2055668319862151.
\bibitem{Mohebbi2020} A. Mohebbi, Human-robot interaction in rehabilitation and assistance: a review, in Current Robotics Reports, 1(3), 2020, pp. 131-144.
\bibitem{Rosati2010} G. Rosati, The place of robotics in post-stroke rehabilitation, in Expert Review of Medical Devices, 7(6), 2010, pp. 753-758.
\bibitem{Billard2007} A. Billard, B. Robins, J. Nadel, J., and K. Dautenhahn, Building Robota, a mini-humanoid robot for the rehabilitation of children with autism, in Assistive Technology, 19(1), 2007, pp. 37-49.
\bibitem{Saleh2021} M.A. Saleh, F.A. Hanapiah, and H. Hashim, Robot applications for autism: a comprehensive review, in Disability and Rehabilitation: Assistive Technology, 16(6), 2021, pp. 580-602.
\bibitem{Wykowska2020} A. Wykowska, Social robots to test flexibility of human social cognition, in International Journal of Social Robotics, 12(6), 2020.
\bibitem{Wykowska2021} A. Wykowska, Robots as mirrors of the human mind, in Current Directions in Psychological Science, 30(1), 2021, pp. 34-40.
\bibitem{Stower2021} R. Stower, N. Calvo-Barajas, G. Castellano, and A. Kappas, A meta-analysis on children’s trust in social robots, in International Journal of Social Robotics, 13(8), 2021, pp. 1979-2001.
\bibitem{García-Corretjer2022} M. García-Corretjer, R. Ros, R. Mallol, and D. Miralles, Empathy as an engaging strategy in social robotics: a pilot study, in User Modeling and User-Adapted Interaction, 2022, pp. 1-39.
\bibitem{Vinanzi2021} S. Vinanzi, A. Cangelosi, and C. Goerick, C., The collaborative mind: intention reading and trust in human-robot interaction, in Iscience, 24(2), 2021, p. e102130.
\bibitem{Dziergwa2018} M. Dziergwa, M. Kaczmarek, P. Kaczmarek, J. Kędzierski, and K. Wadas-Szydłowska, Long-term cohabitation with a social robot: a case study of the influence of human attachment patterns, in International Journal of Social Robotics, 10(1), 2018, pp. 163-176.
\bibitem{Rincon2019} J.A. Rincon, A. Costa, P. Novais, V. Julian, and C. Carrascosa, A new emotional robot assistant that facilitates human interaction and persuasion, in Knowledge and Information Systems, 60(1), 2019.
\bibitem{Rios-Martinez2015} J. Rios-Martinez, A. Spalanzani, and C. Laugier, From proxemics theory to socially-aware navigation: A survey, in International Journal of Social Robotics, 7(2), 2015, pp. 137-153.
\bibitem{Reddy2021} A.K. Reddy, V. Malviya, and R. Kala, Social cues in the autonomous navigation of indoor mobile robots, in International Journal of Social Robotics, 13(6), 2021, pp. 1335-1358.
\bibitem{Tolgyessy2017} M. Tölgyessy, M. Dekan, F. Duchoň, J. Rodina, and P. Hubinský, Foundations of visual linear human–robot interaction via pointing gesture navigation, in International Journal of Social Robotics, 9(4), 2017, pp. 509-523.
\bibitem{Fitzpatrick2008} P. Fitzpatrick, G. Metta, and L. Natale Towards Long-Lived Robot Genes, in Robotics and Autonomous Systems, 56(1), 2008, pp. 29-45.
\bibitem{Cao2019} Z. Cao, G. Hidalgo, T. Simon, S.E. Wei, and Y. Sheikh, Openpose: realtime multi-person 2d pose estimation using part affinity fields, in IEEE transactions on pattern analysis and machine intelligence, 43, 2019, pp. 172–186.
\bibitem{Schroff2015} F. Schroff, D. Kalenichenko, and J. Philbin, Facenet: A unified embedding for face recognition and clustering, in Proceedings of the IEEE conference on computer vision and pattern recognition (CVPR), 2015, pp. 815-823.
\bibitem{Lombardi2022} M. Lombardi, E. Maiettini, D. De Tommaso, A. Wykowska, and L. Natale, Toward an attentive robotic architecture: Learning-based mutual gaze estimation in Human–Robot Interaction, in Frontiers in Robotics and AI, 9, 2022.
\bibitem{He2017} K. He, G. Gkioxari, P. Dollár and R. Girshick, Mask r-cnn, IEEE International Conference on Computer Vision, 2017.
\bibitem{He2016} K. He, X. Zhang, S. Ren, and J. Sun, Deep residual learning for image recognition, in Proceedings of the IEEE conference on computer vision and pattern recognition, 2016
\bibitem{Rudi2017} A. Rudi, L. Carratino, and L. Rosasco. FALKON: An optimal large scale kernel method, in Advances in Neural Information Processing Systems 30, pages 3888–3898, 2017.
\bibitem{Pasquale2016} G. Pasquale, T. Mar, C. Ciliberto, L. Rosasco, and L. Natale, Enabling depth-driven visual attention on the icub humanoid robot: Instructions for use and new perspectives,” Frontiers in Robotics and AI, 2016.
\bibitem{Calli2015} B. Calli, A. Singh, A. Walsman, S. Srinivasa, P. Abbeel, and A.M. Dollar, The ycb object and model set: Towards common benchmarks for manipulation research, in IEEE International conference on advanced robotics (ICAR), 2015, pp. 510-517.
\bibitem{Hampali2020} S. Hampali, M. Rad, M. Oberweger, and V. Lepetit, V., HOnnotate: A method for 3d annotation of hand and object poses, in the IEEE/CVF conference on computer vision and pattern recognition (CVPR), 2020.
\bibitem{COCO} T.Y. Lin, M. Maire, S. Belongie, J. Hays, P. Perona, D. Ramanan, P. Dollár, and C.L. Zitnick, Microsoft coco: Common objects in context, in European conference on computer vision (ECCV), 2014.
\bibitem{Maiettini2020_thesis} E. Maiettini, From Constraints to Opportunities: Efficient Object Detection Learning for Humanoid Robots, Ph.D. thesis, University of Genoa, 2020.
\bibitem{Maiettini2019b} E. Maiettini, G. Pasquale, V. Tikhanoff, L. Rosasco, and L. Natale, A weakly supervised strategy for learning object detection on a humanoid robot, in IEEE-RAS 19th International Conference on Humanoid Robotics (Humanoids), 2019.

\end{thebibliography}
\end{document}